\documentclass{article}
\usepackage[margin=1in]{geometry}

\usepackage{hyperref}
\usepackage{xcolor,amsmath,amsthm,amssymb}
\usepackage{natbib}
 \setcitestyle{aysep={}} 
 \renewcommand{\cite}{\citep}    
 \newcommand{\namecite}{\citet}  

\definecolor{darkblue}{rgb}{0, 0, 0.5}
\hypersetup{colorlinks=true,citecolor=darkblue, linkcolor=darkblue, urlcolor=darkblue}

\usepackage{pdflscape}

\usepackage{tikz,tikz-qtree}
\usepackage{amsmath}

\newcommand{\tikztree}[2]{\begin{tikzpicture}[baseline={([yshift=-.5ex]current bounding box.center)},vertex/.style={anchor=base}]
\tikzset{frontier/.style={distance from root= #2 pt}}
\tikzset{level 1+/.style={level distance=10pt}}
#1
\end{tikzpicture}}

\newcommand\varpm{\mathbin{\vcenter{\hbox{%
  \oalign{\hfil$\scriptstyle+$\hfil\cr
          \noalign{\kern-.3ex}
          $\scriptscriptstyle({-})$\cr}%
}}}}

\usepackage{authblk}
\title{Opening the black box of language acquisition}
\author[1]{Anna Jon-And\thanks{E-mail: anna.jon-and@su.se.}}
\author[1,2]{Jérôme Michaud}
\affil[1]{Centre for cultural evolution, Department of psychology, Stockholm University}
\affil[2]{Division of mathematics and physics, Mälardalen University}
\date{}                     
\begin{document}


%
%


\date{}


\maketitle

\begin{abstract}
Recent advances in large language models using deep learning techniques have renewed interest on how languages can be learned from data. However, it is unclear whether or how these models represent grammatical information from the learned languages. In addition, the models must be pre-trained on large corpora before they can be used. In this work, we propose an alternative, more transparent and cognitively plausible architecture for learning language. Instead of using deep learning, our approach uses a minimal cognitive architecture based on sequence memory and chunking. The learning mechanism is based on the principles of reinforcement learning. We test our architecture on a number of natural-like toy languages. Results show that the model can learn these artificial languages from scratch and extract grammatical information that supports learning. Our study demonstrates the power of this simple architecture and stresses the importance of sequence memory as a key component of the language learning process. Since other animals do not seem to have a faithful sequence memory, this may explain why only humans have developed complex languages.
\end{abstract}

\section{Introduction}
The remarkable and often human-like linguistic performance of modern large language models (LLMs) using deep learning techniques has drawn attention towards language learning and processing. Among other things, it is discussed what we can learn about human linguistic abilities from studying the mechanisms by which LLMs learn from data and generate output from input. 
It is possible to gain some insight into what kind of information LLMs are able to store at the semantic, syntactic and morphological levels by studying their attentional mechanisms \cite{manning2020emergent,vaswani2017attention,piantadosi2023modern,piantasodi2022meaning}. But the neural networks upon which LLMs rely contain billions to trillions of parameters \cite{piantadosi2023modern,brown2020language} and it is not possible to extract information on whether or how any morpho-syntactic abstractions emerge in these models and, if so, what is their nature. In this sense, LLMs are black boxes that generate impressive results but provide us with very limited cues as to what cognitive mechanisms may underlie human language learning. 

In this article, we propose a minimal cognitive architecture that is not a neural network, but follows a simple and traceable error-correction temporal difference learning algorithm \cite{sutton2018reinforcement}, with the aim of opening up the black box of language learning. The minimal architecture allows us, on the one hand, to discuss what minimal cognitive properties are necessary for language learning and, on the other hand, to follow the process of extracting grammatical information that supports the language learning process. Our approach is inspired by cognitive architecture research in artificial intelligence, that aims at building a unified theory of cognition where few general mechanisms should be able to explain diverse phenomena and characteristics of human cognition \cite{newell1994unified}. In line with recent domain-general artificial intelligence modelling \cite{wiggins2020creativity,van2017linking}, we aim at including as few assumptions as possible to account for the language acquisition process. Our initial assumptions, drawing upon a pilot study for the present work \cite{jon2020minimal}, are that sequence memory and chunking are two key domain-general cognitive mechanisms underlying the human language ability. Our approach implements principles of usage-based language learning, i.e. that language acquisition occurs through language use and through continuous updating of linguistic knowledge encoded as chunks or constructions \cite{bybee2006usage,Tomasello2003}.

The task of our learning model is to identify sentences in an artificial linguistic stream where cues to sentence borders like punctuation or capitalization are removed. However, differently from LLMs, our model learns from very limited amounts of data and learning is local and based on simple principles of updating associative strengths in agreement with humans’ limited working memory and processing capacities. Furthermore, the learning task is grounded in cognitively plausible assumptions since the identification of meaningful units in an incrementally perceived stream is a real goal during language learning. In our model, the learner follows a chunk-and-pass learning principle \cite{christiansen2016now} in which it decides whether to place a boundary or to chunk a newly encountered word to other words kept in the working memory. Our chunking mechanism is hierarchic in that it associates chunks to binary tree structures with the words as leaves. Since chunks retain the order in which words are encountered, they encode both the ability for sequence memory and the ability to chunk. Boundary placement triggers reinforcement, where correct sentence identification is rewarded and wrong sentence identification is negatively reinforced.
Our design differs from most syntactic parsing, where the task is to identify parse trees, dependencies and/or constituents, and not to identify sentence boundaries. Our learning mechanism has access to less information than most unsupervised syntactic parsers, that have full access to boundaries as sentences are parsed one at a time, and that generally have access to entire large corpora for probability calculations \cite{klein2002generative,bod2006all}. Our model thus relies on computationally cheaper principles than statistical learning in general. 

The evaluation of the model is based on the one hand on it's ability to fully learn to identify sentences in artificial languages with varying degrees of complexity, and on the other hand on its ability to identify and re-use chunks that are extracted from exposure to these languages. We use probabilistic context-free grammars to generate languages reflecting common natural language structures. We find that the model is able to efficiently identify informative chunks and re-use them when they occur in new structures, thus in a simplified manner accounting for the emergence of grammar during learning. The limited working memory of our model creates an information bottleneck that lets grammatical structure emerge during the learning process as a way to overcome the combinatorial explosion problem inherent to language acquisition. We also find that the model over time learns to ignore less informative chunks and gives priority to those that contribute more frequently to the correct identification of sentences. In this way, we see how a simple task like identification of sentence borders gives rise to self-organization and consolidation in the emergent grammatical system. 

This paper is organized as follows: Section~\ref{sec:back} provides some background for this study. Section~\ref{sec:Comp} presents the computational framework used, i.e. the framework of Markov Decision Processes \cite{sutton2018reinforcement}. In particular, we discuss the learning task as well as the learning algorithms studied in this paper. Section~\ref{sec:Res} presents the results of our study and discusses the performance of our model on a number of small human-like toy languages. We show that our model successfully learns these languages and provide some interesting insights into how learning is performed. Finally, Section~\ref{sec:Disc} provides a discussion of the implication of our results and outlines how this work could be continued.

\section{Background\label{sec:back}}
\subsection{Divergences in language learning in LLMs and humans}

The advent of LLMs has revolutionized natural language processing, with the neural network based Transformer architecture as their foundation \cite{vaswani2017attention}. Even though neural networks are biologically inspired, LLMs are designed to be useful engineering tools and not to model human cognition, and they exhibit notable disparities in learning mechanisms compared to human language acquisition. The difference in training data size is a critical distinction; LLMs typically leverage datasets approximately 1000 times larger than the linguistic input available to a child \cite{warstadt2022artificial}. Moreover, humans learn and use language in parallel  \cite{bybee2006usage,ellis2015usage,Tomasello2003}, while LLMs are generally pre-trained and no longer learn when they are put in use. This implies that LLMs have access to huge resources during learning, not only in terms of data, but also in terms of memory and processing power. Once trained, LLMs can base their output on very long input sequences, sometimes containing thousands of tokens \cite{beltagy2020longformer,liang2023unleashing,brown2020language}. This differs radically from humans, that are typically able to store somewhere between three and ten meaningful items in their working memory \cite{cowan2001magical,Miller1956}. The chunk-and-pass principle, proposed as a strategy for overcoming this constraint, is theorized to underpin the multilevel representational structure of language \cite{christiansen2016now, christiansen2016creating}.  Principles of local decision-making and memory updating, rooted in the utilization of limited recent information and associations derived from past experiences, have also been implemented in models that successfully account for aspects of language acquisition, change, and emergence \cite{mccauley2019language, baxter2006utterance,Michaud2019, steels2012grounded}.

\subsection{Foundation of the human linguistic capacity: Sequence memory and chunking}

The human language capacity stands out among species for its flexibility and expressive power. A fundamental and yet unsolved question is what kind of cognitive mechanisms enable humans to learn grammar. Theories on cognitive mechanisms underlying human language abilities range from inborn language organization \cite{Chomsky1957,pinker2005faculty,bolhuis2014could,christophe2008bootstrapping,valian2013determiners} to compromises between inborn factors and flexibility \cite{Nowak2002,reali2009evolution,thompson2016culture} to domain-general learning \cite{bybee1985morphology,Tomasello2003,heyes2018cognitive}, where linguistic structure would be emergent and dynamic rather than innate \cite{booij2010construction,goldberg2007constructions,croft2004cognitive,croft2001radical,langacker1987foundations}. The linguistic capacities of modern LLMs, that rely on statistical and distributional learning, contradicts claims that language learning needs to rely on specific inborn properties \cite{piantadosi2023modern}.  

Associative learning is a fundamental mechanism in LLMs that also holds explanatory power for general learning in both humans \cite{heyes2018cognitive} and other animals \cite{wasserman2023resolving, lind2018can, heyes2012s, heyes2012simple, enquist2016power, Bouton2016, haselgrove2016overcoming,enquist2023human}. However, the inability of non-human animals to match the language usage capacity of LLMs and humans calls for identifying unique properties that are present in humans, that may be shared by LLMs but not by other animals. Sequence learning has been pointed out as especially important for human linguistic capacities \cite{heyes2018cognitive,christiansen2009usage,bybee2002phonological,christiansen2003language,christiansen2017more,frank2012hierarchical,cornish2017sequence,udden2012implicit,kolodny2015evolution,kolodny2018evolution} and moreover, sequences constitute the fundamental input of LLMs. At the same time, there is strong empirical support for non-human animals limited capacity to represent the exact order of stimuli \cite{Roberts1976,macdonald1993delayed,Ghirlanda2017a,read2021working,lind2023test}. This suggests faithful sequence representation to be a basic defining feature of human cognition and linguistic abilities \cite{enquist2023human,jonsequence}. These findings call for testing the capacity for language learning of a minimalist associative learning model with a precise but limited sequence memory. 

To enable processing of language or of any kind of sequential information, sequence memory likely needs to be combined with a capacity for chunking, i.e. considering a recurrent sequence of stimuli of flexible length as a unit. Without chunking, it is impossible to faithfully represent a sequence. Chunking is known to be central for human language learning \cite{Tomasello2003,bybee2002phonological,servan1990learning,mccauley2019language} and the aforementioned chunk-and-pass principle has been identified as essential for overcoming memory constraints in online language processing \cite{christiansen2016now}.




\subsection{Language models learning formal languages}
  
In this paper, we use probabilistic context-free grammars as a straightforward way to generate diverse hierarchical sentence structures of the kind we know exist in natural languages. A seemingly similar approach is applied in a body of work where language learning models evaluate hypothetical grammars of formal languages via Bayesian inference \cite{rule2018learning,rule2020child,goodman2008rational,goodman2014concepts,piantadosi2016logical,amalric2017language,planton2021theory,ullman2012theory,ellis2018learning}. The formal languages are composed of sequences generated by rule systems known as artificial grammars. These grammars are organized within a mathematical hierarchy of escalating generative power, commonly referred to as the Chomsky hierarchy \cite{chomsky2002syntactic,chomsky1959certain}. In a recent study a learning model discovers the underlying system in a number of context-free and context-sensitive formal languages through exposure to these languages without any pre-defined hypothesis \cite{yang2022one}. Most of the formal languages in these studies do not, however, resemble natural languages. Context-free languages like e.g. $a^nb^n$ are even impossible for humans to learn without explicit counting if $n$ is high. \namecite{hochmann2008humans} showed that human performance in recognizing such strings more likely relies on distributional regularities than any underlying grammar. It is thus unclear what we can learn about the human language capacity by studying the performance of machines on formal languages generated by artificial grammars. This motivates our choice to limit the linguistic input for our learner to structures that are common in natural languages and not explore different levels in the Chomsky hierarchy.  Furthermore, our approach is more cognitively plausible than the approach generally used in the formal language paradigm, as our model does not form any hypotheses on what grammar underlie the data, but instead it relies on emergent dynamic rules to guide each local decision, without necessarily formulating the whole set of rules explicitly. In addition, the model we evaluate is much more minimalist than previous learning models of similar scope, and strives for cognitive plausibility, including only two basic features, sequence memory and chunking, that are well-studied in humans. 

\section{Computational framework\label{sec:Comp}}
The method used in this paper is that of reinforcement learning applied to a Markov Decision Process (MDP) \cite{sutton2018reinforcement}, that encodes a language learning task. The learner tries to identify sentences by placing boundaries between words and gets positive reinforcement when it succeeds and negative reinforcement when it fails. The model will then be evaluated by looking at learning curves, i.e. the fraction of correct responses from a population of agents, and at the grammar generated by the model. This section provides an overview of the learning task, learning algorithms and evaluation methodology.
\subsection{The learning task}
We start by defining the learning task our cognitive architecture will use. Following  \namecite{sutton2018reinforcement}, we define the task as a MDP. MDPs provide a useful framework for studying learning from interactions to achieve a goal. In this framework, a learner is called an \emph{agent}. In our case, the agent aims to identify sentences in a sequence of words constituting the environment. The sequence of words used here is generated using a probabilistic context free grammar (PCFG), but our model should work on any sequential input with well-defined units. To illustrate the type of input the model receives it is useful to take a natural language example. Consider for example the following three sentences: 
\begin{quote}
\emph{The cat chases the dog. The man loves his girlfriend. The sun shines.}
\end{quote}
Since identifying sentences in this context is too easy due to punctuation and capitalization, we create a raw input by removing  any hint as to the beginning of sentences, for example capital letters and punctuation are removed from the list of words, yielding the following modified input:
\begin{quote}
\emph{the cat chases the dog the man loves his girlfriend the sun shines}
\end{quote}
The information about sentence boundaries is stored in a Boolean sequence encoding whether a word in the raw input is preceded by a boundary or not. This Boolean sequence is used to drive the reinforcement procedure. The aim of the learning agent is then to identify sentences, i.e. sequences of words separated by boundaries and not containing any other boundaries. This task can be seen as a masking task, where sentence boundaries are masked and the aim of the model is to predict where boundaries occur.

\subsubsection{MDP formalism: states, actions, and rewards}
In order to specify this task within the MDP framework, we need to specify 
\begin{enumerate}
    \item how agent represent information, i.e. their state space;
    \item what they can do in a given state, i.e. their action spaces, which depends on their state, and
    \item what consequences their actions have, i.e. the reward structure and transitions between states.
\end{enumerate}
 More specifically, the agent interacts with its environment at discrete time steps, $t=0,1,2,3,\dots$. At each time step $t$, the agent is in a given state $S_t\in \mathcal S$ and takes an action $A_t\in \mathcal A(S_t)$. As a result of that action, the agent receives a reward $R_{t+1}\in \mathcal R \subset \mathbb R$ and finds itself in a new state $S_{t+1}\in \mathcal S$. To encode the behaviour of the agent, we use state-action values $Q(s,a)$ encoding the value of taking action $a \in \mathcal A(s)$, when the agent finds herself in state $s$ and define the policy of an agent $\pi(a|s)$ as the probability to select action $a \in \mathcal A(s)$ when in state $s$.

\subsubsection{Specifying the language learning task as a MDP\label{subsec:MDP}}
\begin{figure}
    \centering
    \includegraphics[width=.9\textwidth]{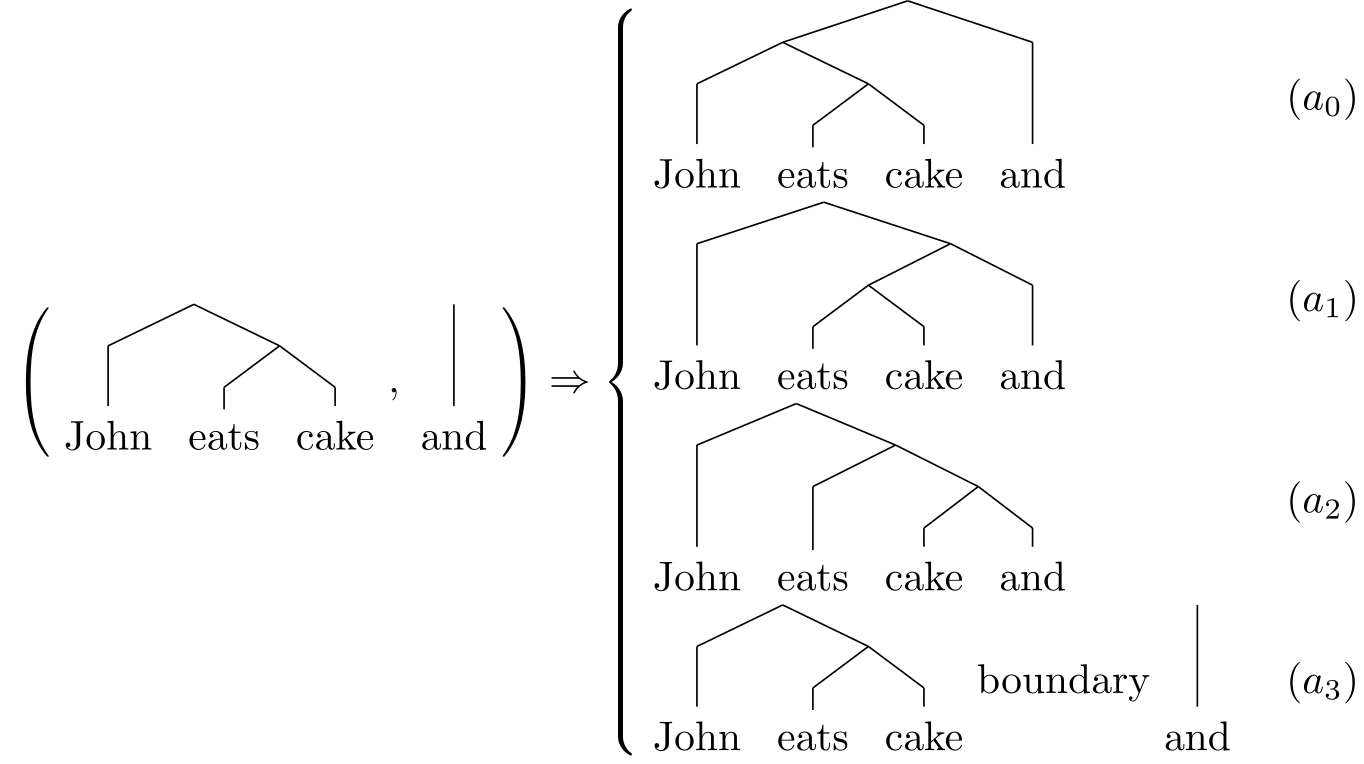}
    \caption{Illustration of the four possible actions. On the right, the results of the four actions are shown. As can be seen, there are three different ways of chunking the second element into the first.}
    \label{fig:chunking}
\end{figure}

In our case, the states are encoded as pairs of structured \emph{chunks} of words. A chunk refers to a sequence of words and an associated binary tree with the words as leaves. By default, the second and most recently perceived chunk always consists of a single word.
An example state is given by:
\begin{equation}S_t = \left(
\tikztree{\Tree  [ [ John ] [ [ eats ] [ cake ] ] ] }{35}
, 
\tikztree{\Tree [ and ] }{35}
\right)\,.\label{eq:state}
\end{equation}
In such a state, the agent has a number of possible actions: she can choose to place a boundary between the two elements of her state, or to integrate the second element into the structure of the first. Figure~~\ref{fig:chunking} illustrates the four possible actions associated with the example above.
The possible actions in a given state depend on the structure of the binary tree and on how many ways it can be chunked with the second element of the state. We define the right-depth of a state $d(S_t)$ as the number of ways the second chunk can be integrated into the first. In the example above $d(S_t)=3$, as is illustrated in Figure~~\ref{fig:chunking}. There are $d(S_t)+1$ possible actions $a_0, \dots, a_{d(S_t)}$ in state $S_t$. We refer to the chunking actions as $a_0$ when chunking at the root, and we increase the index as we go down the tree. The deepest chunking action we can take is $a_{d(S_t)-1}$. For example, action $a_2$ in Figure \ref{fig:chunking} corresponds to grouping the last two stimuli together. The last action $a_{d(S_t)}$ corresponds to placing a boundary between the elements of $S_t$. 

When a chunking action is selected, the resulting chunk becomes the first element of $S_{t+1}$ and the next word is read. In this case, no reward is given.

Upon boundary placement, reinforcement is triggered. Whenever the first element of $S_t$ is a correct sentence, a positive reward of $R_{t+1}=r_+>0$ is given. Otherwise, a negative reward of $R_{t+1}=r_-<0$ is given. The next section describes how the rewards are used to update state-action values and the policy of the agents. After reinforcement, we considered two different ways of reinitializing the task:
\begin{description}
    \item[Continuous:] The second element of $S_t$ becomes the first element of $S_{t+1}$ and the next word is read to instantiate the second element.
    \item[Next sentence:] The first element of $S_{t+1}$ is set to the first word of the next sentence in the sequence of words, and the second element is the following word.
\end{description}
These types of reinitialization process make the MDP episodic in the sense that the process terminates upon boundary placement and restarts in a new state. Thus, each attempt to identify a sentence constitutes a new episode.

\subsection{Learning algorithms}
During learning, an agent must choose between actions in order to increase its expected reward. In this paper, we use a variant of the Q-learning algorithm \cite{sutton2018reinforcement} in which state-action values are updated according to a temporal difference (TD) algorithm. These state-action values can be considered as estimators of the expected reward from that specific state and learning is driven by reducing the error in prediction. In our case, states possess a hierarchic structure whenever $d(S_t)> 1$ and it is possible to modify the standard Q-learning algorithm to take advantage of this structure.

\subsubsection{State-action values, sub-states, and sub-actions}
In order to decide what to do in a given state, an agent relies on the state-action value $Q(s,a)$ of taking action $a$ in state $s$. However, the states we consider here have a hierarchic structure. If the right-depth of a state $d(s_t)> 1$, then there are a number of associated sub-states and corresponding actions. Consider again the state in Eq.~\ref{eq:state}. Since this state has right-depth 3, it has two associated sub-states, namely,
\[S^1_t = \left(
\tikztree{\Tree  [ [ eats ] [ cake ] ]  }{25}
, 
\tikztree{\Tree [ and ] }{25}
\right)\quad \text{and}\quad
S^2_t = \left(
\tikztree{\Tree  [ cake ]   }{22}
, 
\tikztree{\Tree [ and ] }{22}
\right)\,,\]
where we used an upper index to label these sub-states. The original state is labeled $0$ or is left unlabeled. An action $a_i^0 \in \mathcal A_t^0(S^0_t)$ associated with the state corresponds to an action $a_{i-k}^k \in \mathcal A_t^k(S^k_t)$ associated with sub-state $S^k_t$ whenever $i-k \geq 0$. To specify the learning mechanism, it is convenient to use the Heaviside step function $H(x)$ defined as
\[
H(x) = \begin{cases}
    1&\text{ if }x\geq 0\\
    0&\text{ otherwise}
\end{cases}\,.
\]

To take advantage of this structure, we define two composite versions of the state-action values:
\begin{enumerate}
    \item The \emph{average composite state-action value} $\overline Q(s,a_i)$ is defined as:
    \[
\overline Q(s, a_i) := \frac{\sum_{k=0}^{d(s)-1} Q(s^k,a^k_{i-k})H(i-k)}{\sum_{k=0}^{d(s)-1} H(i-k)}\,,\quad i = 0,\dots,d(s)\,;
\]
\item \emph{The additive composite state-action value} $\widetilde Q(s,a_i)$ is defined as:
\[
\widetilde Q(s, a_i) := \sum_{k=0}^{d(s)-1} Q(s^k,a^k_{i-k})H(i-k)\,,\quad i = 0,\dots,d(s)\,;
\]
\end{enumerate}
The average composite state-action value $\overline Q(s,a_i)$ averages the state-action values of a state and its sub-states. The Heaviside function selects the actions that are valid for the sub-states. This quantity will be used in decision making. The additive composite state action value $\widetilde Q(s,a_i)$ computes the sum of state-action values from a state and its associated sub-states. This quantity will be used in the learning mechanism to implement blocking in a similar way as \namecite{Rescorla1972}.

\subsubsection{Updating state-action pairs\label{subsec:algs}}
We are now in position to explain how state-action values are updated. We consider two alternative learning rules:
\begin{description}
    \item[Q-learning:] In this case, state-action values associated with states and sub-states are updated independently. The updating rule is given by
    \[
Q(S^k_t,a^k_{i-k}) \leftarrow Q(S^k_t,a^k_{i-k}) + \alpha \left[R_T - Q(S^k_t,a^k_{i-k})\right]H(i-k)\,.
\]
    \item[Rescorla-Wagner Q-learning:] In this case, state-action values associated with states and sub-states are considered elements of a composite stimulus. The updating rule is given by
\[
Q(S^k_t,a^k_{i-k}) \leftarrow Q(S^k_t,a^k_{i-k}) + \alpha \left[R_T - \widetilde Q(S_t,a_i) \right]H(i-k)\,.
\]
\end{description}
In both cases, $R_T$ is the reward obtained upon boundary placement and $\alpha$ is a learning parameter. The update is performed for all time step $t \in 0,\dots, T$ of the episode and for all sub-states $k\in 0,\dots, d(S_t)$. These TD algorithms are driven by the error in prediction of the reward $R_T$. The Heaviside function is there to avoid updating an action that does not exist in a given sub-state.
The difference between Q-learning and Rescorla-Wagner Q-learning, is that in the first case, the state-action values are used as predictors, whereas in the latter case, we use the additive composite state-action values.

State-action values are initialized by setting 
\[
Q(S_t,a_i) = \begin{cases}
    q_b & \text{if } i=d(S_t),\\
    q_c & \text{otherwise}
\end{cases}\,.
\]
The parameters $q_b$ and $q_c$ encode the initial values of placing a border or choosing a chunking behaviour, respectively.

\subsubsection{Using state-action pairs to make decisions}
In order to choose actions in a given state, $S_t$, we use a soft-max policy. The soft-max policy computes the probability of choosing a given action using the rule
\[
\pi(a_i|S_t)= \frac{e^{\beta \overline Q(S_t,a_i)}}{\sum_{k=0}^{d(S_t)+1}e^{\beta \overline Q(S_t,a_k)}}\,,
\]
where $\beta$ is a parameter controlling the amount of exploration performed by the model. Note that we use the average composite state-action values $\overline Q(S_T,a_i)$ as the support for each action. We want to take advantage of the internal structure of the state and this is the best way to avoid creating biases towards one specific behaviour. We use the same policy for the two possible learning rules mentioned above.
Overall, the model has 6 parameters summarized in Table~\ref{tab:parameters}.
\begin{table}
    \centering
    \begin{tabular}{cl}
         {\bf Parameter}& {\bf Interpretation}\\
         $\alpha$
& Learning rate\\
         $\beta$
& Exploration parameter\\
         $r_+$
& Positive reward\\
         $r_-$
& Negative reward\\
         $q_b$
& Initial value for border\\
         $q_c$& Initial value for chunking\\
    \end{tabular}
    \caption{Parameters of the model.}
    \label{tab:parameters}
\end{table}

\subsection{Evaluation methods}
To evaluate the performance of the model, we need to assess whether it successfully learns the input languages and, if so, how quickly. In addition, we are interested in what grammatical information it extracts. To achieve these goals, we use learning curves to test the ability of the model to learn and quantify its speed. Grammatical information is extracted directly from the state-action values.
\subsubsection{Extracting learning curves}
In order to get a learning curve, we simulate $N$ agents and record for each episode of the MDP whether the decision was correct or incorrect. We then compute the fraction of agents being successful at each episode and plot the fraction of correct responses as a function of the number of episodes of the MDP that we refer to as \emph{trials}. If all agents eventually learn the language perfectly, then the fraction of correct responses should go to one. 

In order to estimate the learning speed, we fit a logistic curve of the form
\begin{equation}\label{eq:logistic}
f(x) = \frac1{1+e^{-k(x-x_0)}}\,,
\end{equation}
that has the two parameters $k$ and $x_0$. $k$ is a measure of the stiffness of the logistic curve and $x_0$ corresponds to the parameter where the function value is $\frac12$. We estimate the learning time as $2x_0$, since at the beginning of the simulation the fraction of correct responses is $0$.
\subsubsection{Extracting grammatical information}
In our model, all grammatical information is encoded in the state-action values of states and their associated sub-states, i.e. in the $Q$ values. In order to extract grammatical information, we need to extract for every state the most likely action to be taken. To do so, we choose a threshold $0<t_G<r_+$ and extract all state-action values greater than $t_G$ and interpret them as grammatical information. Of course, state-action with lower values also constitute grammatical information, but a threshold is necessary to select data to display. For each state, i.e. for each pairs of chunks, we then have the most likely action to be taken and its associated value. This representation of grammatical information does not constitute a standard grammar, but is instructive for understanding how an agent learns. The extracted grammatical information is then saved in an Microsoft Excel spreadsheet. The code of this project as well as all Excel files used in this project are available online in the GitHub repository: \url{https://github.com/michaudj/LanguageLearner/}.

Since one of the aim of this paper is to open the black box of language acquisition, we can use this technique of extracting grammatical information at various stages of the learning process and save them as different sheets in the Excel file. This allows us to look at how the grammatical information of an agent changes over time and whether it converges to a systematic representation.

For more complex languages in which the number of relevant state-action values is too high, we will provide a more qualitative discussion of the grammatical information learned.

\section{Results\label{sec:Res}}
In this section, we present the results of our experiments on various artificial languages. We start by considering a simple language consisting of nouns (N) and verbs (V) organized in sentences with the structure noun-verb-noun (NVN), and then move on to consider more complex grammars. 
\subsection{A baseline case: NVN languages}
Our baseline case has been chosen to illustrate how the agent learns the language and will be used to motivate the choice of learning algorithm used for more complex languages. 
We start by considering a simple NVN language. Our language is generated by a PCFG $G = (\mathcal M,\mathcal T,\mathcal R,\mathcal S,\mathcal P)$, where $\mathcal M$ is the set of non-terminal symbols, $\mathcal T$ is the set of terminal symbols, $\mathcal R$ is the set of production rules, $\mathcal S$ is the start symbol, and $\mathcal P$ is the set of probabilities on production rules. 
\subsubsection{Defining the grammar}
For our simple NVN grammar, we use 
\[
\begin{aligned}
    \mathcal M &= \{S,N,V\}\,,\\
    \mathcal T &= \{n_1,n_2,\dots, n_{K_n}\} \cup \{ v_1,v_2,\dots,v_{K_v}\}\,,\\
    \mathcal S &= S\,,
\end{aligned}
\]
where, $K_n$ is the total number of nouns and $K_v$ is the total number of verbs. The production rules are:
\begin{equation}
R =\left\{ 
    \begin{aligned}
        S & \to  N, V, N\\
        N & \to  n_1|\ n_2|\ n_3| \dots\\
        V & \to  v_1|\ v_2|\ v_3| \dots
    \end{aligned}\right\}
\end{equation}
and are equiprobable. This means that $S$ always rewrites as $N,V,N$, and the nouns $N$ rewrites as $n_i$, $i\in [1,2,\dots, K_n]$ uniformly at random, and similarly for the verbs.

\subsubsection{Learning curves}
\begin{figure}
    \centering
    \includegraphics[width=.9\textwidth]{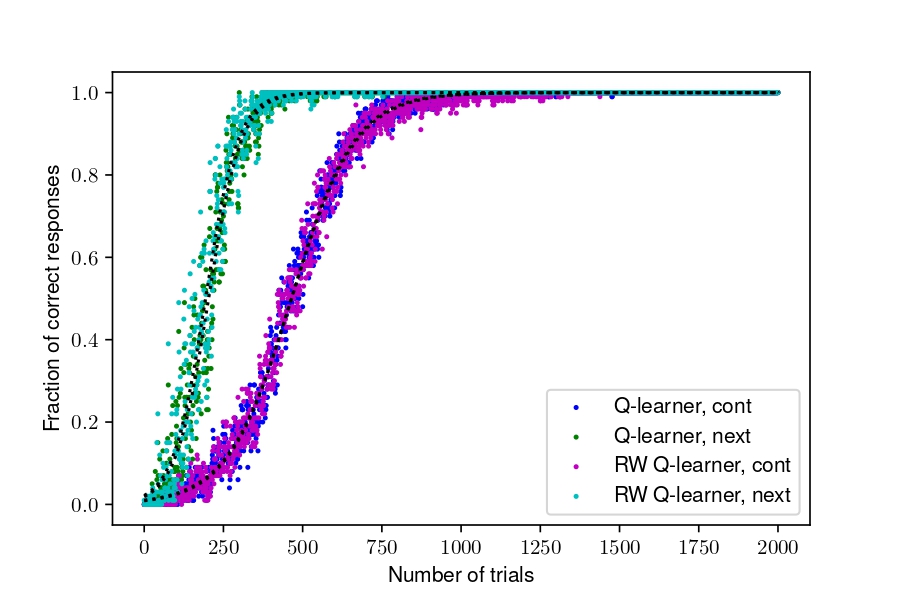}
    \caption{Learning curves for the four combinations of sentence conditions QC, QN, RWQC, and RWQN. The NVN language has $K_n = K_v = 5$. Fractions are obtained over 100 agents.}
    \label{fig:learning_curvesNVN}
\end{figure}

With this NVN language defined, we test the ability of our models to learn this language using the proposed task. We consider the four cases combining the definition of Sections~\ref{subsec:MDP} and \ref{subsec:algs}:
\begin{description}
    \item[QC:] Q-learning with continuous border condition;
    \item[QN:] Q-learning with next sentence border condition;
    \item[RWQC:] Rescorla-Wagner Q-learning with continuous border condition;
    \item[RWQN:] Rescorla-Wagner Q-learning with next sentence condition.
\end{description}
The value of parameters is the same for all models and provided in column "value NVN" of Table~\ref{tab:param_lc}.
\begin{table}
    \centering
    \begin{tabular}{ccccc}
         {\bf Parameter}& {\bf Value} NVN &{\bf Value} MD & {\bf Value} RelClause&{\bf Value} ComplexNP\\
         $\alpha$
& 0.1 &0.1 
 & 0.1 
&0.1 
\\
         $\beta$
& 1.9 &1.0 &  
 1.0 & 1.0 \\
         $r_+$
& 25 &25 
 & 25 
&25 
\\
         $r_-$
& -10 &-10 
 &  -10 
& -10 
\\
         $q_b$
& 1 &1 
 &  
 1 
& 1 
\\
         $q_c$& -1 &-1  &  -1  & -1  \\
    \end{tabular}
    \caption{Parameter values for the different languages studied.}
    \label{tab:param_lc}
\end{table}
\begin{table}
    \centering
    \begin{tabular}{cc}
         {\bf Algorithm}& {\bf Learning time}\\
         QC
& 927\\
         QN
& 402\\
         RWQC
& 925\\
         RWQN
& 388\\

    \end{tabular}
    \caption{Learning times of the four learning algorithms for a NVN language with $K_n=K_v=5$.}
    \label{tab:learning_time}
\end{table}
The obtained learning curves are displayed in Fig~\ref{fig:learning_curvesNVN}. 
We see that all variants of the model successfully learn the language reaching a fraction of correct responses very close to 1. Using the fit to the logistic curve in Eq.~\eqref{eq:logistic}, we estimated the learning times of the four models in Table~\ref{tab:learning_time}.

We observe that when the next sentence condition is used, learning is faster. This is not surprising, since the task is easier because only one of the two boundaries of a sentence has to be identified. With respect to learning times, there is very little difference between Q-learning and Rescorla-Wagner Q-learning, which is expected since the two learning algorithms are very similar. The main difference between these two algorithms is the information used to drive error-correction, i.e. whether blocking is used or not. This will be examined below when we look at what is learned at different stages of the learning process.

\subsubsection{Grammar extraction and acquisition process}
As we have seen above, the performance of Q-learning and Rescorla-Wagner Q-learning are very similar in terms of learning curve. The question addressed in this section is what type of grammatical information is extracted and used by these two learning algorithms and whether the sentence condition affects what is learned. To achieve this,  we report the grammatical information extracted in terms of the two elements of a state/sub-state, the action with the highest state-action value. The results are shown in Tables~\ref{tab:grammarNVNQ} and \ref{tab:grammarNVNQRW} for the continuous border condition and are extracted from the two Excel spreadsheets QLearnerC\_paper.xlsx and RWQLearnerC\_paper.xlsx available in the GitHub repository: \url{https://github.com/michaudj/LanguageLearner/}. Snapshots are taken at the beginning of the S-shaped learning curve, when learning fraction is close to 0.5, when learning is just completed, and at two later points, when performances are very high. The columns labeled \# refer to the number of states $(s_1,s_2)$ matching the pattern. For example, two states $([n_2,v_3],n_4)$ and $([n_1,v_5],n_2)$ have the same structure and constitute two instances of the $([N,V],N)$ state. The results for the next sentence condition are similar and can be found in Excel spreasheets QLearnerN\_paper.xlsx and RWQLearnerN\_paper.xlsx in the GitHub repository.

Our results show that the Rescorla-Wagner algorithm is much more parsimonious in the number of state-action values it relies upon. This is best seen by looking at the number of instances of a given state-action pair reported in columns "\#" of Tables~\ref{tab:grammarNVNQ} and  \ref{tab:grammarNVNQRW}. While these numbers steadily increase in Q-learning, only some of them increase in Rescorla-Wagner Q-learning and some of them even decrease after some time. For example, in the last three rows of Table~\ref{tab:grammarNVNQRW}, the number of instances decreases and even reached 0 for one of them. This shows that the model learns more specific information than the Q-learning algorithm. The same can be observed in the averaged state-action values, where all actions reported increase in value for the Q-learning algorithm, while only some of them increase for Rescorla-Wagner Q-learning and some even decrease after some point. 

The results for Q-learning suggest that all state-action pairs will eventually reach the maximum reinforcement value of 25 and the rule will not specify a preferable tree structure over the sentence. For instance, both $[N, [V, N]]$ and $[[N, V], N]$ tree structure will occur with significant probability. This grammatical output is not very informative.

The results for Rescorla-Wagner Q-learning suggest that if learning continues only 4 state-action pairs are relevant for the decision making process:
\begin{enumerate}
    \item Place border between $N$ and $N$;
    \item Chunk together $N$ and $V$;
    \item Chunk together $V$ and $N$;
    \item When $[N, V]$ is followed by $N$, chunk at the deepest level, yielding $[N, [V, N]]$.
\end{enumerate}

This means that all information about sentence boundary placement is encoded in the $N-N$ transitions and rules 2, 3, and 4 control the parsing process. This model seems much more informative about the parsing process and is, in some sense, minimalist with regard to the information it extracts. Therefore, we choose Rescorla-Wagner Q-learning for the analysis of more complex sentence structures.

\subsection{Towards natural language input}
In this section, we will explore how the Rescorla-Wagner Q-learning algorithm with continuous border condition performs on more complex grammars. We will consider a number of different grammars. First, we consider a grammar that has both mono- and ditransitive verbs to show that when $N-N$ transitions are unreliable to predict sentence transitions, then the model is still able to perform well. We then increase the complexity of the grammars by either introducing relative clauses or a number of other adnominal elements. For these more complex languages, the grammatical information extracted will often be too big to analyze in a table as we have done for the NVN language. Instead, we study the learning curve and break it down by sentence length to get some insight into which sentences are learned first. We then examine some examples of parsed sentences to show some systematic reuse of identified structures. 

\subsubsection{Mono and ditransitive verb}
We first consider a grammar with both monotransitive verbs (MV), that are always followed by one noun, and ditransitive verbs (DV), that are always followed by two nouns. The grammar is defined as follows:  
\[
\begin{aligned}
    \mathcal M &= \{S,N,MV,DV\}\,,\\
    \mathcal T &= \{n_1,n_2,\dots, n_{K_n}\} \cup \{ mv_1,mv_2,\dots,mv_{K_m}\}\cup \{ dv_1,dv_2,\dots,dv_{K_d}\}\,,\\
    \mathcal S &= S\,,
\end{aligned}
\]
where, $K_n$ is the total number of nouns, $K_m$ is the total number of monotransitive verbs, and $K_d$ is the total number of ditransitive verbs. The production rules are
\begin{equation}
\mathcal R =\left\{ 
    \begin{aligned}
        S & \to  N, MV, N|\ N, DV, N, N\\
        N & \to  n_1|\ n_2|\ n_3| \dots\\
        MV & \to  mv_1|\ mv_2|\ mv_3| \dots\\
        DV & \to  dv_1|\ dv_2|\ dv_3| \dots
    \end{aligned}\right\}
\end{equation}
and are equiprobable. This means that $S$ always rewrites as $N, MV, N$ in 50\% of the cases and as $N, DV, N, N$ otherwise. The different word classes rewrites uniformly as in the NVN language above. We refer to this language as the MD language.

\begin{landscape}
\begin{table}
    \centering
    \begin{tabular}{ccc|cc|cc|cc|cc|cc|c}
         $s_1$&  $s_2$&  action & $Q_{300}$&\#& $Q_{600}$&\#&$Q_{900}$&\#&$Q_{2000}$&\#&$Q_{4000}$& \# &Total number\\
         \hline
         $N$&  $N$&  $a_1$ (border) & 6.00&3&11.56&24&20.70&25&24.95&25& 25.00&25 &25\\
         $N$&  $V$&  $a_0$ (chunk) & 6.37&2&11.56&23&20.45&25&24.94&25& 25.00&25 &25\\
         $V$&  $N$&  $a_0$ (chunk) & 5.39&1&11.82&14&19.73&18&24.36&19& 24.90&19 &25\\
         {$[N, V]$}&  $N$&  $a_0$ (chunk at root) & 6.05&1&7.19&13&9.03&61&17.53&87& 23.49&87 &125\\
        {$[N, V]$}& $N$& $a_1$ (chunk deep) & ---&---&6.24&12&8.69&31&17.72&38&23.62&38 &125\\
 {$[V, N]$}& $N$& $a_2$ (border) & 5.56&1&7.21&26&9.41&70&16.76&95& 22.38&95 &125\\
 {$[[N, V], N]$}& $N$& $a_2$ (border) & ---&--- &5.80&8&6.09&40&7.49&150& 11.61&189 &625\\
 {$[N, [V, N]]$}& $N$& $a_3$ (border) & 5.56&1&5.86&13&5.56&76&7.64&318& 11.55&427 &625\\
    \end{tabular}
    \caption{Grammar extracted by the Q-learning algorithm with continuous border. Snapshots are taken after 300, 600, 900, 2000, and 4000 trials. The language is a NVN language with  $R_n=R_v=5$ and we used a threshold of $t_G = 5$. Columns labeled "$Q_i$" report averaged values of $Q((s_1,s_2),a_j)$ over the number of relevant instances (reported in the \# columns) after $i$ trials. The total number column reports the maximum number of instances of the corresponding state-action pairs.}
    \label{tab:grammarNVNQ}
\end{table}

\begin{table}
    \centering
    \begin{tabular}{ccc|cc|cc|cc|cc|cc|c}
         $s_1$&  $s_2$&  action&$Q_{300}$&\#&$Q_{600}$&\#&$Q_{900}$&\#&$Q_{2000}$&\#& $Q_{4000}$& \# &Total number
\\
         \hline
         $N$&  $N$&  $a_1$ (border)&5.27&1&8.48&9&14.80&24&20.60&25& 21.33 &25 &25
\\
         $N$&  $V$&  $a_0$ (chunk)&---&---&6.92&12&16.45&25&24.89&25& 25.00&25 &25
\\
         $V$&  $N$&  $a_0$ (chunk)&---&---&7.75&5&13.15&14&19.01&17& 19.14 &17 &25
\\
         {$[N, V]$}&  $N$&  $a_0$ (chunk at root)&---&---&6.66&4&6.65&13&6.96&19& 8.57 &10 &125
\\
        {$[N, V]$}& $N$& $a_1$ (chunk deep)&---&---&6.31&4&8.30&34&17.25&56&23.14 &56 &125
\\
        {$[V, N]$}& $N$& $a_2$ (border) & ---&---&6.13&4&6.91&21&5.68&10&---&--- &125
\\
 {$[[N, V], N]$}& $N$& $a_2$ (border)&---&---&---&---&6.07&9&5.67&9& 5.50&5 &625
\\
 {$[N, [V, N]]$}& $N$& $a_3$ (border) & ---&---&---&---&5.75&7&5.97&4& 5.36&2 &625\\
    \end{tabular}
    \caption{Grammar extracted by the RW Q-learning algorithm with continuous border. Snapshots are taken after 300, 600, 900, 2000, and 4000 trials. The language is a NVN language with  $R_n=R_v=5$ and we used a threshold of $t_G = 5$. Columns labeled "$Q_i$" report averaged values of $Q((s_1,s_2),a_j)$ over the number of relevant instances (reported in the \# columns) after $i$ trials. The total number column reports the maximum number of instances of the corresponding state-action pairs. }
    \label{tab:grammarNVNQRW}
\end{table}
\end{landscape}

Here, we report results for a MD language with $K_n =5$, and  $K_m = K_d = 1$. We simulate the learning using the Rescorla-Wagner Q-learning with continuous border condition and parameters listed in Table~\ref{tab:param_lc} in the "value MD" column. 

\begin{figure}
    \centering
    \includegraphics[width=.9\textwidth]{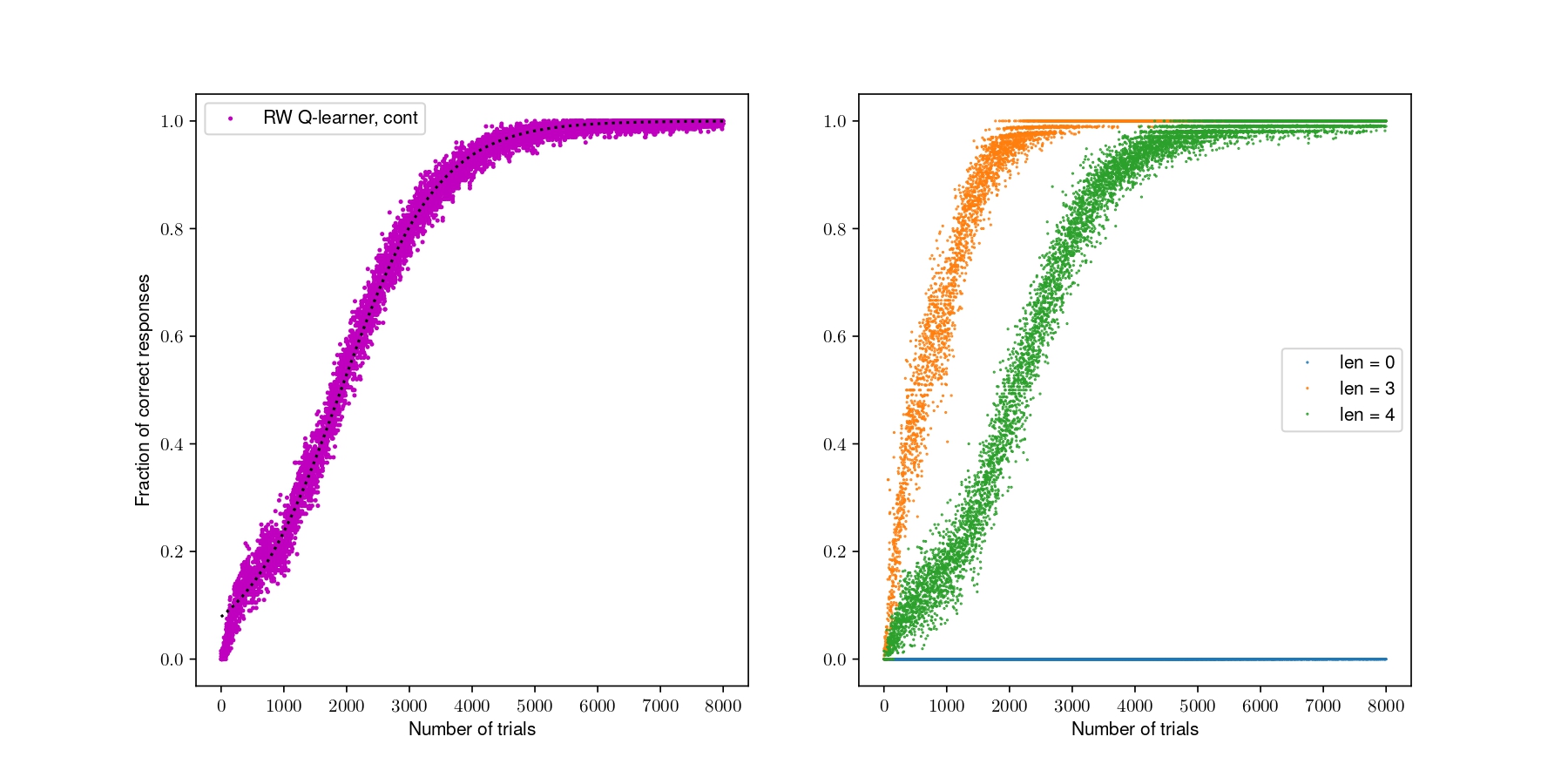}
    \caption{First panel: Learning curve for the Rescorla-Wagner Q-learning with continuous border condition for the MD language with $K_n=5$, $K_m=K_d=1$. Fractions are obtained over 200 agents. Second panel, breakdown of the learning curve by sentence length. Note that if at a given trial no learner encounter a sentence of a given length, then it contributes to sentences of length 0, which are, therefore, meaningless.}
    \label{fig:learning_curveMD}
\end{figure}

Figure~\ref{fig:learning_curveMD} reports the learning curve for this language. The first panel displays the standard learning curve and the second panel shows the learning curve for different sentence lengths. Sentences of length 3 are monotransitive sentences and sentences of length 4 are ditransitive sentences. As can be seen, even when $N-N$ transitions are not a reliable predictor of sentence boundaries, the learning algorithm still learns to identify sentences. The second panel shows that monotransitive sentences are learned first and that there is some interference between the mono- and ditransitive sentences. At the beginning, monotransitive sentences are quickly learned, but the learning slows down as the learner figures out how to deal with ditransitive sentences. After a transition period, both types of sentences are learned. In order to better understand how the processing is done, we look at snapshots taken during the learning process. All data is available online in the Excel spreadsheet RWQLearnerCMD\_paper.xlsx stored in the GitHub repository: \url{https://github.com/michaudj/LanguageLearner/}. The results are displayed in Table~\ref{tab:grammarMDQ}.

The first two snapshots show that at this stage of learning, only information about monotransitive verbs has been learned and decision making relies a lot on full sentences. However, from trial $3000$, the learner starts to rely more on shorter chunks in the decision making process. At this point, the learner starts to have knowledge about ditransitive verbs. At the next snapshot, learning is complete and we see that now boundary placement essentially relies on $N-N$ transitions as in the $NVN$ case. The ditransitive case is solved by using longer chunks containing ditransitive verbs to inhibit border placement and favor a chunking action instead. But in both sentence structures, border placement relies on $N-N$ transitions. This is best seen in the last snapshot where only the $N-N$ transition supports border placements (except for one instance of monotransitive verb, but this is marginal and can be seen as the left over of the previously used chunking strategy).
\begin{landscape}
\begin{table}
    \centering
    \begin{tabular}{ccc|cc|cc|cc|cc|cc|c}
         $s_1$&  $s_2$&  action & $Q_{1000}$&\#& $Q_{1500}$&\#&$Q_{3000}$&\#&$Q_{6000}$&\#&$Q_{25000}$& \# &Total number\\
         \hline
         $N$&  $N$&  $a_1$ (border)& ---&---&---&---&11.45&6&16.30&25& 18.27&25 &25\\
         $N$&  $MV$&  $a_0$ (chunk)& ---&---&---&---&24.65&5&24.99&5& 25.00&5
 &5\\
 $N$& $DV$& $a_0$ (chunk)& ---& ---& ---& ---& 12.40& 4& 24.79& 5& 25.00&5 &5\\
         $MV$&  $N$&  $a_0$ (chunk)& ---&---&---&---&15.85&4&16.04&4& 16.11&4 &5\\
 $DV$& $N$& $a_0$ (chunk)& ---& ---& ---& ---& 12.32& 3& 19.48& 5& 19.55&5 &5\\
         {$[N, MV]$}&  $N$&  $a_0$ (chunk at root)& ---&---&11.22&2&11.81&5&12.06&5& 11.60&6 &25\\
 {$[N, MV]$}& $N$& $a_1$ (chunk deep)& ---& ---& ---& ---& 21.47& 11& 24.99& 11& 25.00&11 &25\\
 {$[N, DV]$}& $N$& $a_0$ (chunk at root)& ---& ---& ---& ---& 10.51& 3& ---& ---& ---&
--- &25\\
 {$[N, DV]$}& $N$& $a_1$ (chunk deep)& ---& ---& ---& ---& ---& ---& 23.80& 1& 25.00&1 &25\\
 {$[MV, N]$}& $N$& $a_2$ (border)& ---& ---& ---& ---& 13.25& 13& 10.19& 1& ---&--- &25\\
 {$[DV, N]$}& $N$& $a_0$ (chunk at root)& ---&---&---&---&10.27&3&14.84&6& 15.33&6 &25\\
 {$[DV, N]$}& $N$& $a_1$ (chunk deep)& ---& ---& ---& ---& 18.31& 2& 20.39& 15& 19.95&16 &25\\
 {$[[N, MV], N]$}& $N$& $a_2$ (border)& 10.64&1&12.44&4&13.14&30&11.63&26& 14.01&1 &125\\
 {$[[N, DV], N]$}& $N$& $a_1$ (chunk deep)& ---& ---& ---& ---& 11.18& 1& 15.90& 8& 19.85&9 &125\\
 {$[N, [MV, N]]$}& $N$& $a_3$ (border)& ---&---&10.93&2&11.33&5&---&---& ---&--- &125
\\
  {$[N, [DV, N]]$}& $N$& $a_2$ (chunk deepest)& ---&---&---&---&10.65&3&18.18&30& 24.98&33 &125
\\
 {$[N, [DV, N]]$}& $N$& $a_1$ (chunk deep)& ---& ---& ---& ---& ---& ---& ---& ---& 13.68&1 &125\\
    \end{tabular}
    \caption{Grammar extracted by the RW Q-learning algorithm with continuous border. Snapshots are taken after 1000, 1500, 3000, 6000, and 25000 trials. The language is a MD language with  $R_n=5$, and $R_m = R_d=1$ and we used a threshold of $t_G = 10$. Columns labeled "$Q_i$" report averaged values of $Q((s_1,s_2),a_j)$ over the number of relevant instances (reported in the \# columns) after $i$ trials. The total number column reports the maximum number of instances of the corresponding state-action pairs.}
    \label{tab:grammarMDQ}
\end{table}
\end{landscape}

\subsubsection{Adding relative clauses}
The next sentences structure we analyze is relative clauses. We add possible relative clauses to the MD grammar above, by letting some nouns be followed by a relative pronoun (Rel), a monotransitive verb and a noun. We choose to only include monotransitive verbs in relative clauses to avoid a combinatorial explosion in the output. The grammar is defined as follows:
\[
\begin{aligned}
    \mathcal M &= \{S,VP,Rel,N,MV,DV,R\}\,,\\
    \mathcal T &= \{n_1,n_2,\dots, n_{K_n}\} \cup \{ mv_1,mv_2,\dots,mv_{K_m}\}\cup \{ dv_1,dv_2,\dots,dv_{K_d}\}\cup \{ r_1,r_2,\dots,r_{K_r}\}\,,\\
    \mathcal S &= S\,,
\end{aligned}
\]
where, $K_n$ is the total number of nouns, $K_m$ is the total number of monotransitive verbs,  $K_d$ is the total number of ditransitive verbs, and $K_r$ is the number of relative pronouns. The production rules are:
\begin{equation}
\mathcal R =\left\{ 
    \begin{aligned}
        S & \to  N, VP\\
        VP &\to MV, N|\ DV, N, N|\ MV, N, Rel|\ DV, N, N, Rel\\
        Rel &\to R, MV, N\\
        N & \to  n_1|\ n_2|\ n_3| \dots\\
        MV & \to  mv_1|\ mv_2|\ mv_3| \dots\\
        DV & \to  dv_1|\ dv_2|\ dv_3| \dots\\
        R & \to  r_1|\ r_2|\ r_3| \dots
    \end{aligned}\right\}
\end{equation}
and the probabilities of the different production rules for the verb phrase $VP$ are set to 0.3 for production rules without relative clause and to 0.2 for production rules with relative clause, i.e. $N, MV, N$ occur in 30\% of the time, $N, DV, N, N$ in 30\%, $N, MV, N, R, MV, N$ in 20\%, and $N, DV, N, N, R, MV, N$ in 20\% of the time. Other production rules are equiprobable. We refer to this language as the RelClause language.

Here, we report results for a RelClause language with $R_n=5$, $R_m=R_d=R_r =1$. Once again, learning is done using the Rescorla-Wagner Q-learning algorithm with continuous border condition. Parameters are the same as in the MD case and displayed in Table~\ref{tab:param_lc}.
\begin{figure}
    \centering
    \includegraphics[width=.9\textwidth]{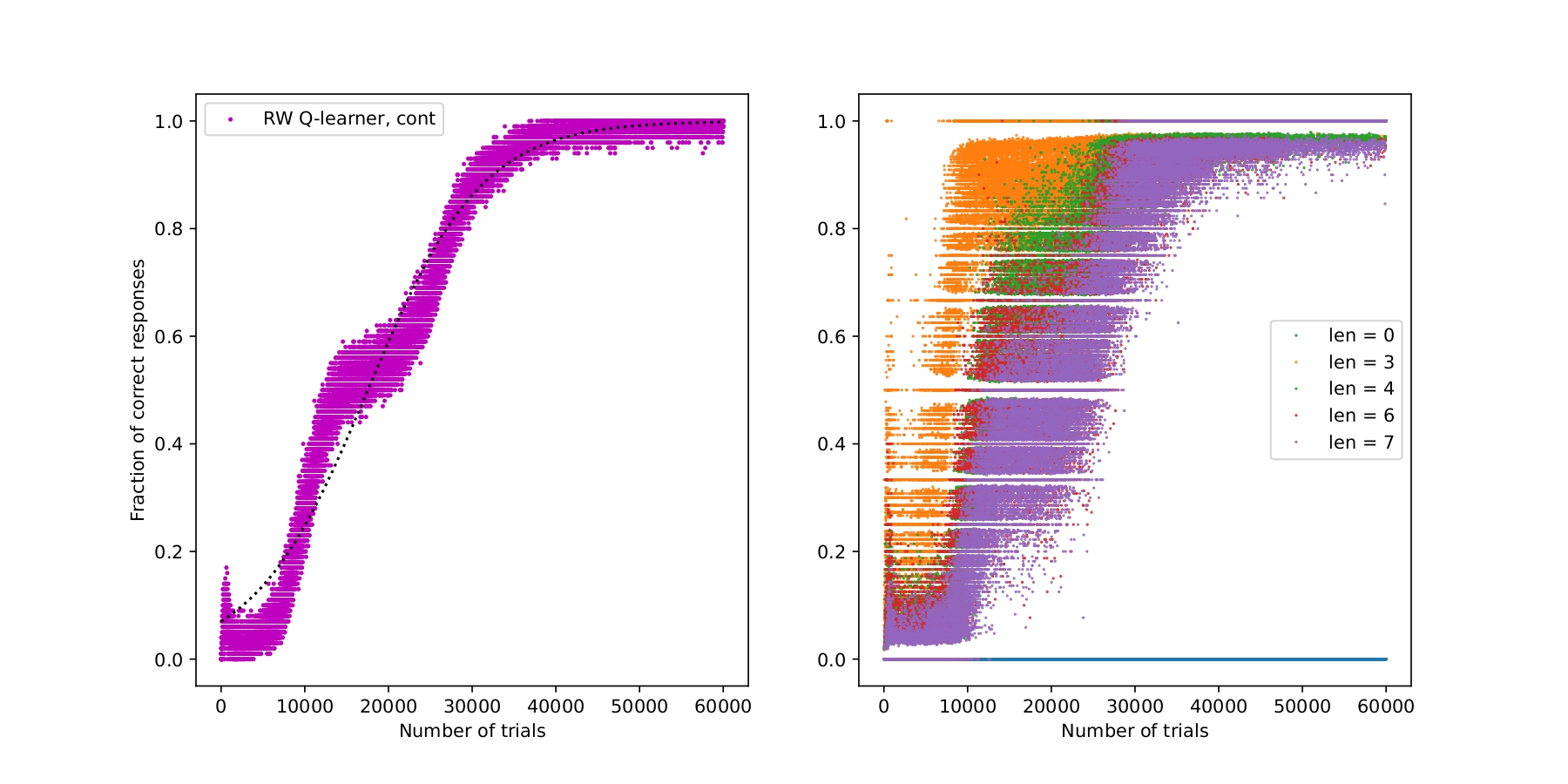}
    \caption{First panel: Learning curve for the Rescorla-Wagner Q-learning with continuous border condition for the relative clause language with $K_n=5$, $K_m=K_d=K_r=1$. Fractions are obtained over 100 agents. Second panel: breakdown of the learning curve by sentence length. Note that if at a given trial no learner encounter a sentence of a given length, then it contributes to sentences of length 0, which are, therefore, meaningless. Since only a fraction of learners encounter a sentence of the same length, explaining the layering structure of the breakdown plot.}
    \label{fig:learning_curveRelClauses}
\end{figure}
The learning curves are displayed in Figure~\ref{fig:learning_curveRelClauses}. Once again, we see that the model learns this language without problems. Interestingly, we observe a U-shaped curve at the beginning of the learning process and we can clearly see that the learning process is irregular and shows a staircase pattern with a plateau around trial 15 000. The breakdown into sentence lengths is showing that at first short sentences are learned well, but then performances decrease while the system learns to deal with ditransitive structures and relative clauses. After some delay all longer sentences get learned without significant delays, pointing out the generalization process over the end of sentences, for instance, 75\% of the sentences end with $MV, N$, making $[MV, N] -N$ transitions a good predictor of sentence boundaries. 

We now take a closer look at the development of the grammatical information of a single learner by extracting snapshots of its state-action values. Selected results are displayed in Table~\ref{tab:grammarRelQ}. Not all constructions are displayed due to their high number. The display is limited to $s_1$ constructions of a maximum length of  $3$ to avoid a combinatorial explosion in the table.  The display is also limited to to constructions that are relatively frequent or that reach relatively high Q-values. The threshold of the the Q-values of the displayed structure is set to $tG = 4$, lower than previous thresholds, in order to capture what is going on in the initial U-curve. For full data on all the state-action values above 4 for this learner are available in the Excel file RWQLearnerC\_rel.xlsx available online in the GitHub repository \url{https://github.com/michaudj/LanguageLearner/}.

In the snapshots we first explore the initial U-shaped curve. We see that in the peak at $400$ trials, the model has learned to identify one $N, MV, N$ sentence by making two chunkings, and then using both the $N -N$ transition and the transition between the full sentence and a following $N$ to place a border. Soon, at $1000$ trials, the model has unlearned to chunk $N -MV$ and to place border between $N -N$, probably because it using the same strategies in longer still unknown sentences and fails. It then relies mainly on the full sentence, that has become difficult to find it when the shorter chunkings are less likely to happen. The shorter transitions are actually very efficient and reliable when the learner is more proficient, and we see that they reappear later, but only around the plateau at $15000$ trials, when the model is starting to learn to deal with relative clauses. At $700$ trials the U-curve has turned and we see that the model starts to learn to handle ditransitive sentences, but it has still not learned to chunk the relative pronoun with preceding constructions. This implies that it still fails to identify the sentences of length $6$ and $7$ and is therefore not able to rely on the shorter constructions that appear in all sentences. At $15000$ trials, the model has started to identify the relative constructions and this is important, because this allows for relying on the shorter transitions to a higher degree. This pattern is strengthened and consolidated at $50000$ trials when the model has learned to correctly identify the sentences with relative clauses and thus performs well on the entire language. Once the transition to the relative clauses is learned, the model learns to recognize the sentences of length $6$ and $7$ more or less in parallel, as the mono- and ditransitive constructions are already learned.

As for the full longer sentences, their tree-structures tend to follow some common patterns at the end of the learning process. The strongest tendency is to deep chunk all verbs with the following noun in both main and subordinate clauses, as in the monotransitive example $[N, [MV, [[N,R], [MV, N]]]]$ or the ditransitive example $[N, [[[DV, N], N], [R, [MV, N]]]]$. This is an efficient re-use of a recurring structure. Because verbs can be preceded by both nouns and relative pronouns, it is more efficient to rely only on the verb for chunking with the following noun than to rely on a chunk that includes one or more elements that precede the verb. We see in the snapshots that single verbs as predictors of chunkings with nouns emerge together with the acquisition of relative constructions at the plateau at $15000$ trials. Just like with the other shorter constructions, their efficiency as predictors depend to a certain extent on the learners ability to identify relative clauses and thus successfully use and re-use them in the longer sentences. The relative pronouns, on the other hand are always precede by a noun and followed by a monotransitive verb in this language. This make the strategies to group them with the preceding noun, as in the first example,  or with the following verb, as in the latter example, equally efficient. This results in a lot of variation between these strategies. 

\begin{landscape}
\begin{table}
    \centering
    \begin{tabular}{ccc|cc|cc|cc|cc|cc|c}
         $s_1$&  $s_2$&  action & $Q_{400}$&\#& $Q_{1000}$&\#&$Q_{7000}$&\#&$Q_{15000}$&\#&$Q_{50000}$& \# &Total number\\
         \hline
         $N$&  $N$
&  $a_1$ (border)& 4.81&1&---&---&---&---&14.15&25& 16.87&25&25
\\
         $N$&  $MV$
&  $a_0$ (chunk)& 6.22&1&---&---&---&---&24.58&5& 24.53&5&5
\\
 $N$& $DV$
& $a_0$ (chunk)& ---& ---& ---& ---& ---& ---& 16.00& 5& 24.94&5&5
\\
 $N$& $R$
& $a_0$ (chunk)& ---& ---& ---& ---& ---& ---& 12.56& 4& 16.82& 4&5
\\
         $MV$&  $N$
&  $a_0$ (chunk)& ---&---&---&---&---&---&19.43&5& 23.05&5&5
\\
 $DV$& $N$
& $a_0$ (chunk)& ---& ---& ---& ---& ---& ---& 18.59& 5& 17.43&5&5
\\
 $R$& $MV$
& $a_0$ (chunk)& ---& ---& ---& ---& ---& ---& 20.20& 1& 22.47& 1&1
\\
        {$[N, MV]$}&  $N$
&  $a_0$ (chunk at root)& ---&---&---&---&5.61&6&5.35&14& ---&---&25
\\
 {$[N, MV]$}& $N$
& $a_1$ (chunk deep)& 7.80& 1& ---& ---& ---& ---& 4.02& 1& 25.00&1&25
\\
 {$[N, DV]$}& $N$
& $a_0$ (chunk at root)& ---& ---& ---& ---& 5.24& 1& 7.87& 23& 7.48&
25&25
\\
 {$[N,N]$}& $N$
& $a_2$ (border)& ---& ---& ---& ---& 4.25& 1& 5.95& 43& 4.51&22&125
\\
 {$[R,MV]$}& $N$
& $a_1$ (chunk deep)& ---& ---& ---& ---& ---& ---& ---& ---& 23.13&1&5
\\
 {$[MV, N]$}& $N$
& $a_2$ (border)& ---& ---& 4.06& 1& 4.75& 4& 10.06& 25& 5.81& 22&25
\\
 {$[MV,N]$}& $R$
& $a_0$ (chunk at root)& ---&---&---&---&---&---&10.59&4& 8.00&2&5
\\
 {$[MV,N]$}& $R$
& $a_1$ (chunk deep)& ---& ---& ---& ---&--- &--- & 7.21& 1& 20.54&1&5
\\
 {$[DV, N]$}& $N$
& $a_0$ (chunk at root)&--- &---&---&---&---&---&15.55&15& 18.14&15&25
\\
 {$[DV, N]$}& $N$
& $a_0$ (chunk deep)&--- &---&---&---&---&---&16.63&9& 18.58&11&25
\\
 {$[[N, MV], N]$}& $N$
& $a_2$ (border)& 5.28&1&5.33&7&7.35&22&8.18&25& 6.68&25&125
\\
 {$[N, [MV, N]]$}& $N$
& $a_3$ (border)&--- &--- &--- &--- & 6.87& 16& 7.15& 84&--- &---&125
\\
 {$[N, [MV, N]]$}& $R$
& $a_1$ (chunk deep)&--- &---&---&---&---&---&5.44&3& 4.46&5&25
\\
 {$[N, [MV, N]]$}& $R$
& $a_2$ (chunk deepest)&--- &---&---&---&---&---&6.97&1& 25.00&2&125
\\
 {$[N, [DV, N]]$}& $N$
& $a_1$ (chunk deep)&--- &---&---&---&---&---&8.14&25& 5.52&43&125
\\
 {$[N, [DV, N]]$}& $N$
& $a_2$ (chunk deepest)&--- &---&---&---&---&---&8.08&18& 25.00&19&125
\\
 {$[MV,[N,R]]$}& $MV$
& $a_1$ (chunk deep)&--- &---&---&---&---&---&---&---& 13.36&2&5
\\
 {$[MV,[N,R]]$}& $MV$
& $a_2$ (chunk deepest)&--- &--- &--- &--- &--- &--- &--- &--- & 20.90&1&5
\\
{$[[DV,N],N]$}& $N$
& $a_3$ (border)&--- &--- &--- &--- & 5.10& 3&--- &--- & 5.53&57&125
\\
{$[[DV,N],N]$}& $R$
& $a_1$ (chunk deep)&--- &--- &--- &--- &--- &--- &--- &--- & 20.00&2&25
\\
{$[DV,[N,N]]$}& $N$
& $a_3$ (border)&--- &--- &--- &--- & 4.25& 1& 7.08& 46& 4.60&25&125
\\
{$[DV,[N,N]]$}& $R$
& $a_2$ (chunk deepest)&--- &--- &--- &--- &--- &--- & 5.38& 4& 21.00&5&25
\\
    \end{tabular}
    \caption{Grammar extracted by the RW Q-learning algorithm with continuous border. Snapshots are taken after 400, 1000, 7000, 15000, and 50000 trials. The language is a MD language with  $R_n=5$, and $R_m = R_d=1$ and we used a threshold of $t_G = 4$. Columns labeled "$Q_i$" report averaged values of $Q((s_1,s_2),a_j)$ over the number of relevant instances (reported in the \# columns) after $i$ trials. The total number column reports the maximum number of instances of the corresponding state-action pairs.}
    \label{tab:grammarRelQ}
\end{table}
\end{landscape}

\subsubsection{Complexifying NPs}
In this section, instead of adding relative clauses to the noun phrase (NP), we add andvary several shorter adnominal elements see how the model handles these more complex and variable NPs. We add determiners (D) and one or two adjectives (A) before the noun, and prepositional phrases consisting of a preposition (P) and a noun after the noun. The language we use is specified by:
\[
\begin{aligned}
    \mathcal M &= \{S,NP,VP,N,MV,DV,A,D,P\}\,,\\
    \mathcal T &= \{n_1,n_2,\dots, n_{K_n}\} \cup \{ mv_1,mv_2,\dots,mv_{K_m}\}\cup \{ dv_1,dv_2,\dots,dv_{K_v}\}\cup \{ a_1,a_2,\dots,a_{K_r}\}\\
    &\cup \{a_1,a_2,\dots, a_{K_a}\} \cup \{ d_1,d_2,\dots,d_{K_d}\}\cup \{ p_1,p_2,\dots,p_{K_p}\}\,,\\
    \mathcal S &= S\,,
\end{aligned}
\]
where, $K_n$ is the total number of nouns, $K_m$ is the total number of monotransitive verbs,  $K_v$ is the total number of ditransitive verbs, $K_a$ is the number of adjectives, $K_d$ is the number of determiners, and $K_p$ is the number of prepositions. The production rules are
\begin{equation}
\mathcal R =\left\{ 
    \begin{aligned}
        S & \to  NP, VP\\
        NP & \to N|\ D, N|\ D, A, N|\ D, A, A, N|\ N, P, N\\
        VP & \to MV, NP|\ DV, NP, NP\\
        N & \to  n_1|\ n_2|\ n_3| \dots\\
        MV & \to  mv_1|\ mv_2|\ mv_3| \dots\\
        DV & \to  dv_1|\ dv_2|\ dv_3| \dots\\
        A & \to  a_1|\ a_2|\ a_3| \dots\\
        D & \to  d_1|\ d_2|\ d_3| \dots\\
        P & \to  p_1|\ p_2|\ p_3| \dots
    \end{aligned}\right\}\,,
\end{equation}
where all productions rules are equiprobable except for NPs which rewrite according to the following probabilities: 
\begin{itemize}
    \item $NP \to N$: 0.25
    \item $NP \to D,N$: 0.25
    \item $NP\to D, A, N$: 0.1875
    \item $NP \to D, A, A, N$: 0.0625
    \item $NP\to N, P, N$: 0.25
\end{itemize}
This means that double adjectives are less common than single adjective NPs, and NP with adjectives are as common as other NP production rules. We refer to this language as a ComplexNP language.

\begin{figure}
    \centering
    \includegraphics[width=.9\textwidth]{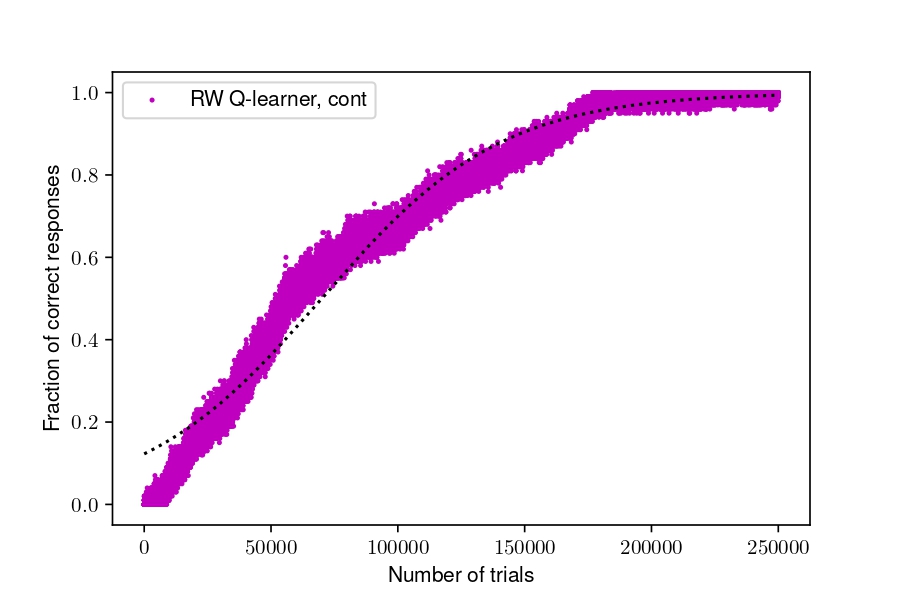}
    \caption{Learning curve for the Rescorla-Wagner Q-learning with continuous border condition for the ComplexNP language with $K_n=K_m=K_v=K_a=K_d=K_p=1$. Fractions are obtained over 100 agents.}
    \label{fig:learning_curveComplexNP}
\end{figure}

Here, we report result for a ComplexNP language with $K_n=K_m=K_v=K_a=K_d=K_p =1$. This language contains 150 different sentence structures, which is much more than the 4 possibilities of the RelClause language above. This is why we restrict the analysis to a smaller vocabulary. Figure~\ref{fig:learning_curveComplexNP} displays the learning curve for that language. We again used the Rescorla-Wagner Q-learning algorithm with continuous border condition. As we can see, the learner successfully learns this language but as in the RelClause language, the learning curve does not follow a standard S-shaped curve. The fit to the logistic is thus imperfect to capture the learning time. We chose not to display the breakdown of the learning curve by sentence length because many different sentence structures are of the same length, for example, compare the sentences $D, N, MV, N$, to $N, DV, N, N$. Both have length 4, and there are other structures of length 4, such as $N, MV, D, N$. This ambiguity is even stronger for longer sentences, which justifies only displaying the overall learning curve.

Figure~\ref{fig:learning_curveComplexNP} is characterized by an irregular learning curve, where short plateaus can be seen. The learning curve displays a period of quick learning following by a period of slow learning. Why this pattern occurs is unclear, but to get an idea of how the sentences are processed, we study one specific learner and record the sentences correctly identified between trial 500 000 and  trial 1 000 000, i.e. after learning is complete. Due to the large number of possible sentences, we only study 5 sentence structures, those starting with $D, A, N, DV, D, A, N$, i.e. ditransitive sentences in which the first two NPs rewrite as $D,A,N$. Data for these 5 sentences as well as other structures is available in the Excel spreadsheet sentencesCompNP\_paper.xlsx in the GitHub repository \url{https://github.com/michaudj/LanguageLearner/}. For these five sentence structures, we report the frequency of each tree structure in Table~\ref{tab:ComplexNP}. Looking at the most frequent structures, we observe that they are very similar. Compare the most likely structure for the first and third sentences in the table. The structures are $[[D, A], [[N, [DV, [[D, A], N]]], N]]$ and $[[D, A], [[N, [DV, [[D, A], N]]], [[N, P], N]]]$, which only differ in the parse of the last NP where $N$ is replaced by $[[N, P], N]$. Similar analysis holds for the three other sentences, which shows the high reuse of extracted information.
\begin{table}
\small
    \centering
    \begin{tabular}{ccc}
         Sentence&  Tree structure& Frequency\\
         \hline
         {$D, A, N, DV, D, A, N, N$}&  {$[[D, A], [[N, [DV, [[D, A], N]]], N]]$}& 0.62\\
         &  {$[[D, A], [[N, [DV, [D, [A, N]]]], N]]$}& 0.17\\
         &  {$[[[D, A], [N, [[DV, D], [A, N]]]], N]$}& 0.17\\
         &  {$[[D, A], [N, [[DV, D], [[A, N], N]]]]$}& 0.04\\
         
         \hline
         {$D, A, N, DV, D, A, N, D, N$}&  {$[[D, A], [N, [[[DV, [[D, A], N]], D], N]]]$}& 0.55\\
         &  {$[[[D, A], [N, [[DV, [D, [A, N]]], D]]], N]$}& 0.16\\
          & {$[[D, A], [N, [[[DV, D], [A, N]], [D, N]]]]$}&0.10\\
         &  {$[[D, A], [N, [[DV, [[D, A], N]], [D, N]]]]$}& 0.06\\
         &  {$[[D, A], [[N, [[[DV, D], [A, N]], D]], N]]$}& 0.05\\
         & {$[[D, A], [N, [[DV, D], [[A, N], [D, N]]]]]$}&0.03\\
         &  {$[[D, A], [[N, [DV, [[D, A], [N, D]]]], N]]$}& 0.02\\
          & {$[[D, A], [N, [[[DV, D], [[A, N], D]], N]]]$}&0.02\\
         &  {$[[D, A], [N, [[DV, [D, [[A, N], D]]], N]]]$}& 0.01\\
 & {$[[D, A], [N, [[DV, D], [A, [[N, D], N]]]]]$}&0.01\\ 
 \hline
         {$D, A, N, DV, D, A, N, N, P, N$
}&  {$[[D, A], [[N, [DV, [[D, A], N]]], [[N, P], N]]]$}& 0.49\\
& {$[[[D, A], [N, [[DV, D], [A, N]]]], [[N, P], N]]$}&0.18\\
&  {$[[D, A], [[N, [DV, [D, [A, N]]]], [[N, P], N]]]$}& 0.15\\
&  {$[[D, A], [[[N, [DV, [[D, A], N]]], [N, P]], N]]$}& 0.08\\
&  {$[[D, A], [[N, [DV, [[D, A], N]]], [N, [P, N]]]]$}& 0.04\\
&  {$[[D, A], [[N, [[DV, D], [[A, N], N]]], [P, N]]]$}& 0.03\\
&  {$[[D, A], [[N, [DV, [D, [A, N]]]], [N, [P, N]]]]$}& 0.02\\
         &  {$[[D, A], [N, [[DV, D], [[A, N], [[N, P], N]]]]]$}& 0.01\\
 \hline
 {$D, A, N, DV, D, A, N, D, A, N$
}& {$[[D, A], [[N, [[[DV, [[D, A], N]], D], A]], N]]$}&0.56\\
& {$[[D, A], [N, [[[[DV, [D, [A, N]]], D], A], N]]]$}
&0.09\\
& {$[[D, A], [N, [[[[DV, D], [A, N]], D], [A, N]]]]$}
&0.09\\
& {$[[D, A], [N, [[[DV, [D, [A, N]]], D], [A, N]]]]$}&0.07\\
&  {$[[D, A], [N, [[[DV, [[D, A], N]], [D, A]], N]]]$}
& 0.05\\ 
&  {$[[D, A], [N, [[[DV, D], [A, N]], [[D, A], N]]]]$}
& 0.05\\
&  {$[[D, A], [[N, [[DV, D], [[A, N], D]]], [A, N]]]$}
& 0.04\\
&  {$[[[D, A], [N, [DV, [[D, A], [N, D]]]]], [A, N]]$}
& 0.02\\
 & {$[[D, A], [N, [[DV, [[D, A], N]], [[D, A], N]]]]$}
&0.01\\
 & {$[[D, A], [N, [[DV, [D, [A, N]]], [[D, A], N]]]]$}
& 0.01\\       
 & {$[[D, A], [N, [[[DV, D], [A, N]], [D, [A, N]]]]]$}
&0.01\\
\hline
 {$D, A, N, DV, D, A, N, D, A, A, N$
}&  {$[[D, A], [[N, [[[DV, [[D, A], N]], D], [A, A]]], N]]$}
& 0.45\\
&  {$[[D, A], [N, [[[DV, [D, [A, N]]], D], [[A, A], N]]]]$}
& 0.15\\
&  {$[[D, A], [N, [[[[DV, [[D, A], N]], D], A], [A, N]]]]$}
& 0.09\\
&  {$[[D, A], [[N, [[DV, [[D, A], N]], [[D, A], A]]], N]]$}
& 0.06\\
& {$[[D, A], [N, [[[DV, D], [A, N]], [[[D, A], A], N]]]]$}
&0.06\\
 & {$[[D, A], [N, [[[[DV, D], [A, N]], D], [[A, A], N]]]]$}
&0.06\\
& {$[[D, A], [[N, [[DV, D], [[A, N], D]]], [[A, A], N]]]$}
&0.04\\
         &  {$[[D, A], [N, [[[DV, [D, [A, N]]], D], [A, [A, N]]]]]$}
& 0.03\\     
         &  {$[[[D, A], [N, [DV, [[D, A], [N, D]]]]], [[A, A], N]]$}
& 0.02\\
         &  {$[[D, A], [N, [[DV, [D, [A, N]]], [[D, A], [A, N]]]]]$}
& 0.01\\
    \end{tabular}
    \caption{Sample of parses of sentences from the ComplexNP language. We report only the parses of sentences starting by $D,A,N,DV,D,A,N$. The second column shows the possible tree structures and the last column report frequencies between the 500 000 and the 1 000 000 trials}
    \label{tab:ComplexNP}
\end{table}

Another thing we can discuss it the parsimony of the grammatical information extracted. Although there are many different structures available in this table, the number of possible tree structures is actually a lot larger. It is given by the Catalan number $C_{n-1}=\frac1n\binom{2(n-1)}{n-1}$  for a sentence of length $n$. This means that for a sentence of length $8$ such as the first sentence of Table~\ref{tab:ComplexNP} there are $C_7=429$ different tree structures, which is much larger than the 4 listed above. The same argument holds for longer sentences. We have about $10$ tree structures for the other sentences, but the Catalan number increases extremely rapidly. In fact, we have $C_8=1430$, $C_9=4862$, $C_{10}=16796$, which means that the 10 tree structures for the last sentence are among the $16796$ possibilities.
\section{Discussion\label{sec:Disc}}
Our results show that our model is able to identify and re-use chunks of information that facilitates language learning, thus in a simplified manner accounting for the emergence of grammar during learning. For a simple language only containing the sentence structure noun-verb-noun, we see that the model is able to efficiently avoid the combinatorial explosion that comes with recognizing each sentence individually by only relying on noun-noun transitions to identify sentence borders. In a slightly extended language with both mono- and ditransitive verbs,  containing the sentence structures noun-verb-noun and noun-verb-noun-noun, the noun-noun transition is no longer a reliable cue to sentence borders. The model then learns to use the information in the ditransitive verbs to avoid border placement between the nouns following them, and keeps the noun-noun transition as support for border placement in all other cases. The model is also able to learn languages with extended complexity, including relative clauses and noun phrases with determiners, adjectives and prepositional phrases. In these languages, that include many common structures of natural languages, we see how the model gradually learns to identify and re-use sequences of words that often occur together, like verbs and following nouns, or elements of noun phrases, thus reproducing empirical observations of natural language acquisition, where structures are initially acquired by learning frequency-based multi-word sequential chunks \cite{tomasello2005constructing,tomasello2008acquiring,bybee2002sequentiality,arnon2011brush}. This re-use of learned information results in high parsimony in the emergent tree-structures, where only a tiny fraction of all the possible tree structures actually occur. Studying learning curves and snapshots from our models' acquisition of different languages also show that shorter sentences are acquired earlier than longer, similarly to natural language acquisition \cite{hoff2009blackwell}. Furthermore, in la language with relative clauses we see an interesting development where the model initially learns to rely on some frequent noun-verb and noun-noun transitions for identifying sentences, and then over-uses this knowledge before it has identified longer less frequent constructions or the transition to relative clauses. The model thus reproduces a U-shaped curve that can be compared to those resulting from over-regularization of many different kinds that typically occur in early phases of language acquisition \cite{bowerman1982starting,plunkett2020u,marcus1992overregularization,ramscar2013error} 

Another interesting result concerns the use of the blocking mechanism inspired by the Rescorla-Wagner model when learning to respond to a hierarchical structure. It is well-observed that when animals learn to respond to compound stimuli containing one known element that is sufficient for making a good decision, they do not update any stimulus-response associations for the other elements in the compound, thus blocking unnecessary learning about new stimuli \cite{Rescorla1972}. Here we apply the same principle when our model is exposed to a hierarchical structure where one state can contain many dependent sub-states, with possibilities to recruit information from all levels. A state with one or more sub-states can be compared to a compound stimulus. We tested the model without blocking, i.e. reinforcing the state-action values for all levels in the hierarchy separately and with blocking, where all levels are reinforced jointly. We found that blocking leads to a more parsimonious grammar and seems to be a more efficient strategy for extracting relevant information from linguistic input, since the more informative levels are allowed to block the reinforcement of less informative levels. 

Our cognitive architecture uses the framework of  Markov Decision Processes. This framework can be related to memory systems and other cognitive theories. We can interpret the state-action values as representing the long-term memory of the learners. Information about action selection during an episode of the Markov Decision Process \cite{sutton2018reinforcement} constitute the short-term memory. States, sub-states and the selected actions are stored and manipulated during the processing of a sentence and this information is used for decision-making and reinforcement. This processing system resembles what Baddeley calls the Central Executive in his model of the working memory \cite{baddeley2015memory}.  Furthermore, the limited access to information in the working memory is compatible with the chunk-and-pass principle of \namecite{christiansen2016now} and mirrors the memory limitations discussed in  \cite{cowan2001magical,Miller1956}.
All these aspects of our model stresses its cognitive plausibility and highlight its minimalist nature.


The architecture we present and its learning conditions are radically different from those of LLMs, both in terms of the size of the training data and in terms of the length of the strings that are used as input, that are both very limited in our case. Furthermore, our model operationalizes usage-based language learning \cite{bybee2006usage,ellis2015usage,Tomasello2003}, where learning and use occur in parallel, which is not generally the case for LLMs. Our minimal temporal difference model is also different from neural networks in general, as it aims at maximal simplicity. This makes results easier to trace and interpret, and it also enables us to make as few assumptions as possible and see how far they can take us in understanding the human language learning process. Our results indicate that limiting  working memory and training data may be cues to understanding human language learning, as our model learns to find the shortest and most informative chunks as a strategy to overcome these constraints. Our results also indicate that a limited sequence memory combined with chunking, are central components that may be sufficient to learn language.

Our model is limited by the lack of generalization and abstraction, preventing it from scaling to larger languages. We see that learning times increase exponentially when increasing the size of the vocabulary or the complexity of the artificial languages. While our results provide an initial proof of concept that the minimal cognitive architecture we present suffices for extracting grammatical information, future studies should explore generalization to generate abstract categories that would enable testing the model on larger languages and eventually apply it to natural languages. Emergent abstract categories would then be expected to resemble grammatical categories like word classes or syntactic constituents. Error-correction models have previously successfully accounted for the emergence of phones \cite{milin_tucker_divjak_2023} and the learning of regular and irregular morphological forms \cite{ramscar2013error,macwhinney1989language} in a human-like manner from exposure to limited parts of language that are relevant to these tasks. The modeling of the emergence of syntactic categories is arguably a more challenging and more interesting problem, as it needs to involve tracing relationships between words and constituents at different levels, and requires exposure to a somewhat complete language. A generalization system in our model should generalize over the information that the model uses, i.e. information about sequences and chunks, as we have done manually to report the results in this study. One way of implementing this is using the mathematical theory of type systems, being the simplest mathematical framework for handling ordering, composition and abstraction of things \cite{heunen2013quantum}, thus incorporating the features sequence memory, chunking, and schematizing. In generative linguistics, an application of this framework is made in the theory of syntactic types \cite{lambek1958mathematics}. A pilot study, where the theory of syntactic types is modified to be dynamic and compatible with usage-based learning, shows that functional abstract categories can emerge for short sentences \cite{jon2023}, indicating that this is a promising framework for further development.

Another limitation of our model is that it only considers syntax and not semantics or meaning. This complicates the learning task since, meaning of words highly constrain the possible parsing of a sentence.  Form-meaning pairings are considered fundamental for usage-based learning and the compatible framework of construction grammar \cite{langacker2002concept,goldberg2007constructions}. In LLMs, even though words are not mapped onto real-world meanings, meaning can be represented in a distributional sense, and this has been shown to be important for their performance \cite{piantasodi2022meaning,manning2020emergent}. This distributional, text-related sense of meaning could be implemented in a minimalist model like ours and would likely be a key for it to perform well on natural language corpora. A further possible future development of our model would be to use the emergent grammar it abstracts for linguistic production, and to let production and feedback to production be part of the learning process. This would be an important progress towards testing the models ability to account for the complete language acquisition process. There are still quite a few steps that needs to be taken to achieve this, but our initial results motivate further development and exploration of this minimal cognitive architecture.

\section*{Acknowledgments}
We thank our colleagues at the Centre for Cultural Evolution, Stockholm University for all inspiring and helpful discussions. This work was supported by the Swedish Research Council (VR 2022-02737). 
\bibliography{compling_style}

\end{document}